# Khmer Text Classification Using Word Embedding and Neural Networks

Rina Buoy          Nguonly Taing          Sovisal Chenda

Techo Startup Center (TSC)

{`rina.buoy,nguonly.taing,sovisal.chenda`}@techostartup.center

**Abstract**

Text classification is one of the fundamental tasks in natural language processing to label an open-ended text and is useful for various applications such as sentiment analysis. In this paper, we discuss various classification approaches for Khmer text, ranging from a classical TF-IDF algorithm with support vector machine classifier to modern word embedding-based neural network classifiers including linear layer model, recurrent neural network and convolutional neural network. A Khmer word embedding model is trained on a 30-million-Khmer-word corpus to construct word vector representations that are used to train three different neural network classifiers. We evaluate the performance of different approaches on a news article dataset for both multi-class and multi-label text classification tasks. The result suggests that neural network classifiers using a word embedding model consistently outperform the traditional classifier using TF-IDF. The recurrent neural network classifier provides a slightly better result compared to the convolutional network and the linear layer network.

## 1 Introduction

Classification is one of the fundamental features of human and machine intelligence. Recognising words, letters, images and voices are essentially assigning a category to an input, presented to our senses. Classification lies also at the heart of natural language processing (NLP), of which one of the classification tasks is known as text categorization or classification. Text classification is a process of assigning a label or category to an entire text or document. Common applications of text classification include:

- Sentiment analysis: a task involved in extracting sentiment orientation toward some objects such as products or movie reviews [1; 2].

- Spam detection: a binary classification task involved in assigning an email to one of the two classes namely: spam or non-spam [1; 2].

- Article classification: a task involved in classifying a submitted manuscript or article into a relevant topic of the magazine or journal [3].

Text classification can be done via handwritten rules or supervised machine learning. The former is fragile due to changes in data or situations while the latter learns the mapping from an input to a correct output, automatically. Under the supervised machine learning approach, text is commonly represented using a bag-of-words model from which features are extracted and fed to a classifier [1].

In this paper, we apply various supervised learning algorithms to Khmer text classification task. Unlike English or other well-studied languages in which text classification is rather straightforward, Khmer is an unsegmented, low-resource language. Word segmentation, thus, is a required preprocessing step for downstream tasks [4].

After examining limitations of a traditional text feature extractor using a bag-of-words model and classifier on Khmer text, we propose the following solutions to improve classification performance by taking into account Khmer language specifications:

- Khmer word embedding model: instead of a semantic unaware bag-of-

words model, a FastText[1] word embedding model is proposed. The model was trained using subword representations that are able to locate semantically-related words. The model was trained on 1-million-sentence corpus which has approximately 30 million words. The trained model is able to handle out-of-vocabulary words as well as words with spelling errors, thanks to subword representations.

- Neural Network classifiers: instead of using an explicit text feature extractor, end-to-end neural network models are proposed. The models learn implicit feature representations from data during training. The proposed models are linear layer network, recurrent neural network (RNN), and convolutional neural network (CNN).

The following sections of the paper are organized as follows: In section 2, background of Khmer language and related work are provided. In section 3, we elaborate the detailed description of the proposed solutions. Section 4 explains data collection, models setup, and training while section 5 and 6 provide results and discussion, followed by a conclusion in section 7.

## 2 Background

### 2.1 Khmer Script

Khmer (KHM) is the official language of the Kingdom of Cambodia. The Khmer script is used in the writing system of Khmer and other minority languages such Kuay, Tampuan, Jarai Krung, Brao and Kravet. Khmer language and writing system were hugely influenced by Pali and Sanskrit in early history [5; 6].

The Khmer script is an abugida that was descended from the Brahmic script. Unlike most Indo-European languages, Khmer script does not use any explicit word delimiter and the definition of words is not a natural concept [7]. Word segmentation, thus, is a required preprocessing step for downstream tasks [4].

---

[1]https://fasttext.cc/docs/en/crawl-vectors.html

### 2.2 Khmer Text Classification

Text classification is one of the most active research areas in most languages. There are, however, very few published papers on the topic of Khmer text classification at the time of writing. One of the reasons may be due to the fact that there are not many publicly available text classification datasets in Khmer.

[8] published a pipeline for classifying Khmer news articles which consists of the following steps:

1. Word segmentation;
2. Feature extraction using term frequency-inverse document frequency (TF-IDF); and
3. Classifier models training.

Article [8] applied the above pipeline on a binary classification tasking using classifiers such as Naive Bayes (NB), Logistic Regression (LG), Support Vector Machine (SVM), Random Forest (RF), and XGBoost. XGBoost outperformed the rest by achieving an accuracy rate of 98%.

The same author also applied the same methodology to a multi-class text classification task [9]. The highest accuracy rate achieved by XGBoost and logistic regression, was 89% which is 9% lower than the binary classification task.

Article [3] presented a similar methodology for classifying articles in Vietnamese which is also an unsegmented language which does not any explicit word delimiter. The highest accuracy achieved by SVM was around 91.4%.

As a feature extractor, TF-IDF suffers from the following limitations:

- Lack of context: Since TF-IDF is a variant of bag-of-words model, context or word order is not considered. However, Khmer is a highly analytic language in which morphemes can be combined freely to different words or expressions [7].

- No use of semantic similarities between words: TF-IDF does not take into account semantic similarities between words.

- Inability to handle out-of-vocabulary words or spelling mistakes. Spelling mistakes can be caused by word segmentation.

## 2.3 Khmer Word Embedding

There exists a pre-trained word embedding model for Khmer language. The model is a FastText model that is trained on Khmer Wikipedia corpus by using ICU tokenizer as a word segmenter. ICU tokenizer is a dictionary-based approach to identify words. Such approach is sensitive to spelling mistakes, unable to handle out-of-vocabulary (OOV) words, and does not take into account context [10; 4]. Khmer is highly analytic in morphology and, therefore, context is important in identifying token or word boundaries [11].

## 3 Proposed Solutions to Khmer Text Classification Task

We propose the following solutions to Khmer text classification task:

- Subword embedding of Khmer words to capture semantic similarities.

- End-to-end neural network classifiers

The detailed explanations are given in the following sections.

### 3.1 Semantic Modelling of Khmer Words

To capture semantic similarities between words, a word embedding model is proposed. A FastText embedding model uses a subword model by representing each word as itself, plus a combination of sub-words [12]. A FastText embedding model is proposed for the following reasons:

- Representing a word as a series of subwords is at the heart of Khmer character cluster concept [13].

- Computing word embedding from subwords tolerates spelling errors. For example, រមកិរ្ណ៍ and រមកិរ្ណ៍ should have similar embedding vectors as most of subwords are the same except for a missing ់.

Training FastText embedding model is an unsupervised learning task that requires only segmented corpus. The model can be trained into two ways; namely, skipgram and cbow (continuous-bag-of-words). The skipgram model learns representations by predicting a target using a nearby word while cbow predicts a target word by using its context words. Figure 1 and 2 show how cbow and skipgram work for an input sentence - ខ្ញុំ ចូលចិត្ត និយាយ ភាសា ខ្មែរ (I like to speak Khmer).

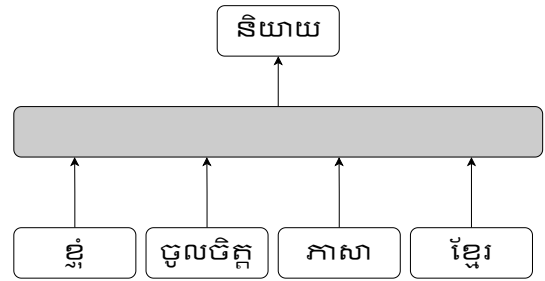

Figure 1. Continuous bag-of-words (cbow)

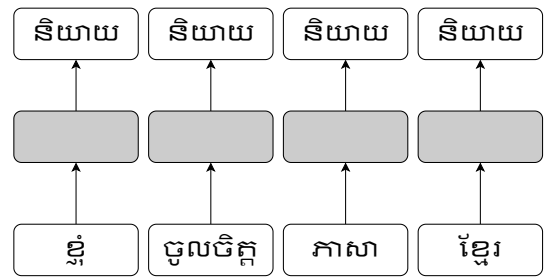

Figure 2. Skipgram

### 3.2 Neural Network Classifiers

#### 3.2.1 Linear Layer Model

A linear layer model for text classification was proposed by [14] as a trick to obtain decent a classification performance while minimizing model size. The key idea of a linear layer model is to take an average of word embedding in a given sentence or text. The average embedding vector is then fed to a linear classifier, which is a realization of neural networks, to predict class probabilities. This is illustrated in Figure 3.

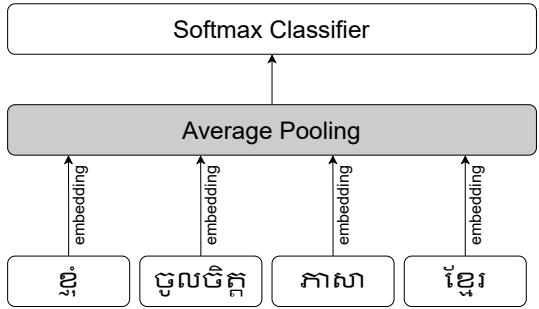

Figure 3. Architecture of a linear layer model

### 3.2.2 Recurrent Neural Network Model (RNN)

Instead of taking an average of word embedding vector, a recurrent neural network model processes each word embedding vector in sequence in both directions via a stack of recurrent units [1]. The recurrent units return an concatenated hidden vector that can be fed to a linear classifier. This is illustrated in Figure 4.

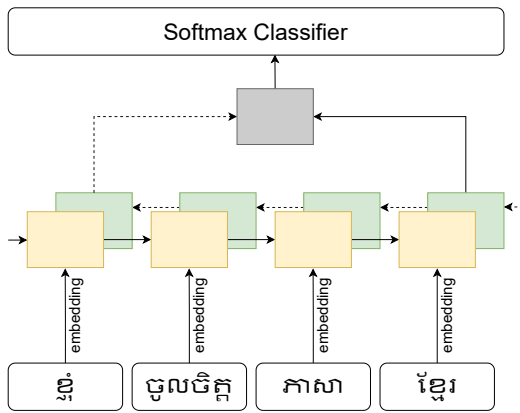

Figure 4. Architecture of a recurrent neural network model

### 3.2.3 Convolutional Neural Network Model (CNN)

A CNN model is traditionally used to analyse images. The model is made up of convolutional layer(s) and linear layer(s). A convolutional layer uses filters or kernels to produce feature maps which are, then, fed to downstream convolutional layer(s) or a linear classifier. A sentence or text is just a collection of words and single dimension in nature. However, when words are represented as embedding vectors, a collection of words becomes 2 dimensions as shown in Figure 5 [15]. A CNN model for text classification is, thus, made up of convolutional blocks followed by max pooling layers and a linear classifier [16]. The model architecture is given in Figure 6.

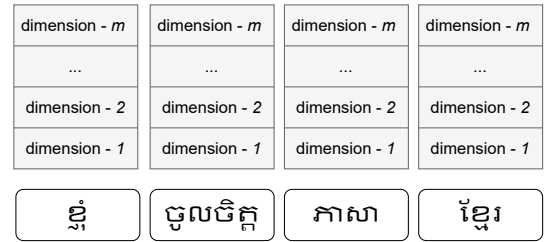

Figure 5. 2D representation of a sentence or text

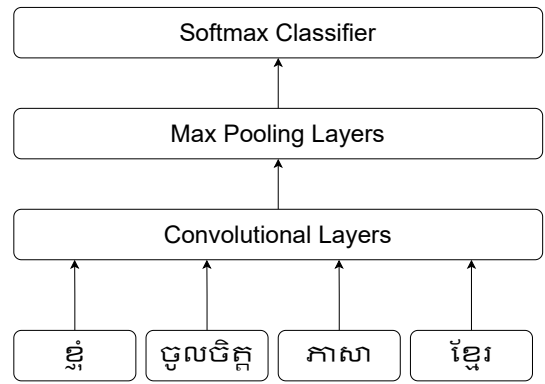

Figure 6. Architecture of a convolutional neural network model

## 4 Data Collection, Models Setup, And Training

### 4.1 Data Collection

The pre-trained Khmer FastText model is not used for the reasons given above. A new Khmer FastText model was trained on a richer and larger dataset. An in-house context-based word segmenter was used to prepare the dataset. The word segmenter is a bidirectional recurrent neural network model that utilizes contextual information from both left and right [4]. Compared with ICU tokenizer, the word segmenter is less prone to out-of-vocabulary words and spelling errors. The primary data sources are Wikipedia texts and online news articles from various sites. The aggregated dataset is made up of approximately 1 million sentences that are about 30 million words.

Text classification data was prepared by collecting news articles from a local news website, Thmey Thmey. Each news article can have one or more labels/tags, assigned by article author(s). The total number of articles is 13902 of which 4687 articles have single label. The data is, thus, split into two smaller datasets; namely, multi-class and multi-label. The multi-class dataset consists of those articles with single label while the multi-label is made of those with one or more labels.

### 4.2 Word Embedding Model

The embedding model was trained by using FastText that is a library for efficient learning of word representations and sentence classification. The model was trained on Google Colab's virtual machine that is Intel(R) Xeon(R) CPU @ 2.30GHz with a ram of 32GB. It took about 12 minutes using FastText's default settings. The result of FastText binary file is approximately 800MB.

### 4.3 Classifier Models Setup

The specifications for the neural network models are shown in Figure 7, Figure 8, and Figure 9 of which the notations are given below:

- $n$: sequence length
- $m$: embedding dimension (100)
- $k$: number of class labels
- $h$: hidden dimension
- $f(.,.)$: filter size
- $f$: number of filters (50)
- $dropout\ prob. = 0.5$

The numbers of trainable parameters are:

- Linear layer model: 21,607
- RNN model: 163,007
- CNN model: 46,207

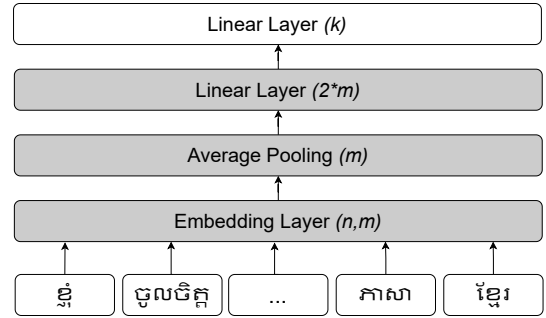

Figure 7. Linear layer model's specifications

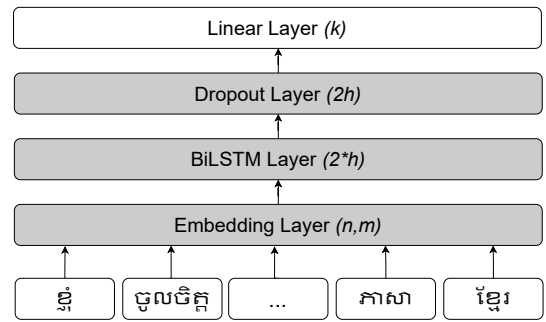

Figure 8. RNN model's specifications

### 4.4 Classifier Models Training

Each classifier is trained on two datasets:

- Multi-class classification: each article is assigned to a single class label and cross-entropy loss function is used.

- Multi-label classification: each article is assigned to one or more class labels and binary cross-entropy loss function is used instead. Since a multi-label classification task is just an extension from multi-class classification task, it is inevitably more difficult [17].

For each case, 500 articles are used as a validation set to evaluate model performance.

## 5 Results

### 5.1 Multi-class Classification

The proposed models were trained on the multi-class dataset and the performance metrics on the test set are reported in Table 1. The SVM classifier with TF-IDF was also trained and used as a baseline model for benchmark.

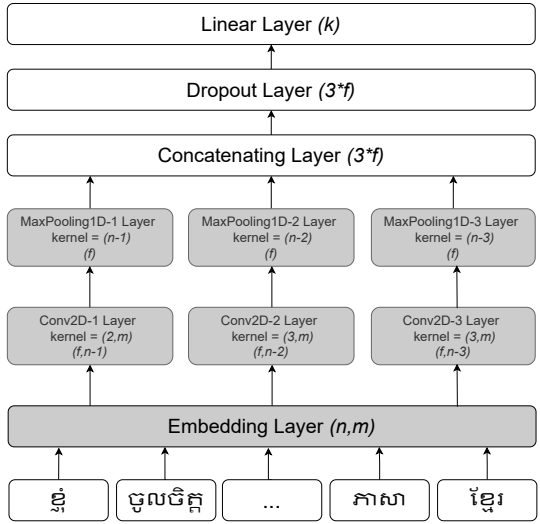

Figure 9. CNN model's specifications

| Models | Precision | Recall | F1 Score |
|---|---|---|---|
| SVM+TF-IDF | 0.84 | 0.84 | 0.84 |
| Linear Model | 0.929 | 0.928 | 0.926 |
| RNN | 0.951 | 0.952 | 0.951 |
| CNN | 0.949 | 0.95 | 0.948 |

Table 1. Performance comparisons for the multi-class classification task on the test set

All of the proposed models using word embedding outperformed the baseline classifier with TF-IDF by a margin of around 10%. The CNN model achieved comparable results as those of the RNN model albeit slightly lower. Both CNN and RNN model outperformed the linear layer model by around 2% across all metrics.

### 5.2 Multi-Label Classification

The performance metrics of all models in the multi-label classification case were consistently lower than the multi-class task as the multi-label classification task is more challenging.

The RNN and CNN model achieved comparable metrics and were the top models, followed by the linear layer model. The proposed models using word embedding consistently outperformed the baseline classifier.

### 6 Discussion

The experimental results for both classification tasks suggest that using word embed-ding improves F1-score up to around 6% from the baseline, while using contextual information (i.e. RNN or CNN model model) provides up to 5% F1-score improvement additionally.

| Models | Precision | Recall | F1-Score |
|---|---|---|---|
| SVM + TF-IDF | 0.812 | 0.73 | 0.761 |
| Linear Model | 0.828 | 0.816 | 0.8127 |
| RNN | 0.878 | 0.855 | 0.862 |
| CNN | 0.888 | 0.836 | 0.853 |

Table 2. Performance comparisons for the multi-label classification task on the test set

The word embedding model implicitly encodes weak contextual information via its context window during training although the learnt representations are static. That means a word can has only one embedded vector regardless of the context. The word embedding model can, nevertheless, significantly improves classification performance metrics.

Figure 10 shows the 2D projections of some words using principal component analysis (PCA), illustrating that semantic similarities between words are reasonably captured. Spelling variants or errors of a word are located close to one another in the embedding space. A word embedding model can, thus, help classifiers to handle spelling errors introduced by word segmentation process.

The RNN and CNN models provide dynamic contextual information which is, otherwise, not captured in static word embedding. Since the RNN model captures context from both directions, it performs slightly better than the CNN counterpart despite of having around 4 times more trainable parameters.

### 7 Conclusion

In this paper, we discussed the challenges and solutions associated with Khmer text classification task. While the task may be rather straightforward in some languages, understanding Khmer language specifications and preprocessing steps is a key to improve classification performance. We presented a Khmer word embedding model that captures semantic similarities between words at subword level. Since the word embedding model uses subwords, it can handle OOVs

Page 5 of 7

Figure 10. 2D projections of some words using principal component analysis (PCA))

and spelling errors, caused by prior segmentation step. We proposed three neural network models that utilise the word embedding model in text classification. The three models are linear layer, recurrent, and convolutional neural networks. The experimental results with multi-class and multi-label classification datasets proved that the neural network models using the word embedding model consistently outperformed the baseline model using TF-IDF by a margin of up to 10% across performance metrics. The RNN model using bidirectional context achieved the highest performance metrics among the 3 proposed neural networks.